\documentclass[11pt,a4paper]{article}
\usepackage[hyperref]{emnlp2018}
\usepackage{times}
\usepackage{latexsym}
\usepackage{graphicx}
\usepackage{url}
\usepackage{CJKutf8}
\usepackage{amsmath}
\usepackage{bm}
\usepackage[group-separator={,},group-minimum-digits={3}]{siunitx}
\usepackage{subfigure}
\usepackage{epsfig}
\usepackage{amssymb}
\usepackage{multirow}

\aclfinalcopy

\title{Identifying High-Quality Chinese News Comments Based on Multi-Target Text Matching Model}

\date{}

\begin{document}
\author{AAAI Press\\
Association for the Advancement of Artificial Intelligence\\
2275 East Bayshore Road, Suite 160\\
Palo Alto, California 94303\\
}

\maketitle

\begin{abstract}
With the development of information technology, there is an explosive growth in the number of online comment concerning news, blogs and so on. The massive comments are overloaded, and often contain some misleading and unwelcome information. Therefore, it is necessary to identify high-quality comments and filter out low-quality comments. In this work, we introduce a novel task: high-quality comment identification (HQCI), which aims to automatically assess the quality of online comments. First, we construct a news comment corpus, which consists of news, comments, and the corresponding quality label. Second, we analyze the dataset, and find the quality of comments can be measured in three aspects: informativeness, consistency, and novelty. Finally, we propose a novel multi-target text matching model, which can measure three aspects by referring to the news and surrounding comments. Experimental results show that our method can outperform various baselines by a large margin on the news dataset.

\end{abstract}

\section{Introduction}
With the development of information technology, more and more people begin to express their opinions on the Internet, leading to an explosive growth in the number of on-line comment concerning news, blogs and so on. These massive comments not only cause information overload, but also contain lots of misleading and unwelcome information. Therefore, it is necessary to identify high-quality comments and filter out low-quality comments. In this paper, we explore how to automatically assess the quality of online comments based on their text data and the relevant auxiliary information, which we call the task of high-quality comment identification (HQCI).

A task similar to the HQCI is text classification. However, general text classification tasks are usually designed based on a single input text, but the quality of the comments is influenced by a variety of factors. For instance, the quality of the comment itself, the consistency of the comment and the corresponding topic, and so on. This leads to a fundamental question: what are the crucial aspects that characterize a high-quality comment? By analyzing of a large number of comments, we find that the key factors affecting the quality of comments lie in the following three aspects:

\begin{itemize}
    \item  \textbf{Informativeness:} A high-quality comment is usually informative and contains sufficient useful information.
    \item \textbf{Consistency:} A high-quality comment is usually highly consistent with the corresponding topic, which is decided by the corresponding news.
	\item \textbf{Novelty:} A high-quality comment tends to be novel, distinguishable and able to stand out from a large number of comments.
\end{itemize}

\begin{CJK*}{UTF8}{gbsn}
\begin{table*}[tb]
\centering
\small

\label{table-example}
\begin{tabular}{l|p{11.0cm}}
\hline
Title    & 国家车辆选号系统遭受黑客攻击。\\ & The national vehicle license plate selection system gets hacked. \\ \hline
Abstract & 出人意料的是黑客攻击了国家车辆选号系统，他们使用这一系统获得了很多有着好的号码的车牌，并且出售这些车牌以牟利。 \\ & It is beyond our imagination that hackers invade the national vehicle license plate selection system. They use the system get many plates of good number, and then sell them for profit.  \\ \hline
Body     & 为什么那些有着好的号码的车牌那么难以获得？选择系统存在着什么问题么？...\\ & Why are those vehicle license plates with good number are hardly to get? Is there anything wrong with the selection system...  \\ \hline
Type & 社会 \quad Society \\ \hline
Comment \#1 & 车辆号牌能够买卖使我觉得搞笑,政策不是规定禁止车牌买卖吗？ \\ & It makes me feel funny that the vehicle license plate can be sold or bought. Isn't it forbidden by the policy? \\ & (247 Likes, 3 Replies)                                  \\ \hline
Comment \#2  &前排抢沙发！ \\ & I am the first one to make a comment! \\ & (0 Likes, 0 Reply) \\ \hline                             
\end{tabular}
\caption{An example of Toutiao Comment Dataset. The original text in the dataset is in Chinese, so we give the translation of the text. And for each comment, we show the likes number and replies number.}
\end{table*}
\vspace{-0.1in}
\end{CJK*}

The measurements for consistency and novelty are about two parts of texts (comment and news, comment and surrounding comment). So in this view, the HQCI can be seen as a subtask of Natural Language Sentence Matching (NLSM). But different from the traditional sentences matching tasks, such as answer selection and paraphrase identification, which usually contain two parts of texts. In HQCI task, we need to consider the matching between comment and different kinds of auxiliary information at the same time. So we propose the Multi-Target Text Matching (MTM) model, which can automatically assess the quality of on-line comments by referring to the relevant auxiliary information including news title, news abstract, and surrounding comments. More specifically, our model measures the informativeness of a comment by the comment itself, the consistency by matching the comments with the news, and the novelty by referring to the surrounding comments. Experimental results show that our model's scoring are highly correlated with human scoring in all of the aspects. 

It is a big challenge for HQCI that we lack annotated dataset for news comments. And we need comments' quality label to conduct supervised method. To overcome this problem, we propose the Toutiao Comment Dataset for this task. It contains the user-generated information that can be used as the quality label of comment. 

The contributions of this paper are listed as follows:
\begin{itemize}
\item We propose the task of high-quality comment identification (HQCI), and construct a large-scale annotated dataset for this task. 
\item We perform human evaluation to analyze the relationship between the quality of comments and three metrics: informativeness, consistency, and novelty. The human evaluation shows that these metrics can measure the quality of comments well.
\item In order to measure three metrics above automatically, we propose Multi-Target Text Matching model (MTM), which can identify high-quality comments by referring to different kinds of auxiliary information. Experimental results show that our model's scoring are highly correlated with human scoring in three aspects. Besides, our model outperforms various baselines by a large margin. 
\end{itemize}

\section{Toutiao Comment Dataset}
\label{dataset}

In this section, we introduce the Toutiao Comment Dataset. The existing comment datasets, such as SFU Opinion and Comments Corpus~\citep{kolhatkar2018sfu-data2}, do not contain the annotated quality information, so they are not suitable for the HQCI task.
%
%
Therefore, we construct Toutiao Comment Dataset, which contains news and comments. More importantly, the dataset contains annotated quality information, i.e. the number of likes, which is naturally generated by users. 

%

\begin{table}[]
\centering
\small
\begin{tabular}{c|ccc}
\hline
\bf Attribute     & \bf Avg-Word & \bf Avg-Char & \bf Vocab \\ \hline
Title    & 16.64     & 24.02     & 36378     \\
Abstract & 75.95     & 114.24    & 46533     \\
Body     & 326.17    & 523.78    & 63425     \\
Comment  & 18.37     & 25.67     & 53916     \\ \hline
\end{tabular}
\caption{Statics information of the textual attributes (Avg-word and Avg-char denote the average number of words and characters, respectively. Vocab means the vocabulary size).}
\label{table-statics}
\vspace{-0.15in}
\end{table}

Table~\ref{table-example} shows an example of our data. Each piece of data has 5 attributes: title, abstract, body, type, and a list of comments, and each comment has associated numbers of likes and replies. Table~\ref{table-example} also shows two examples of comments. It shows that a high-quality comment is more likely to get more likes than a common one, so it is reasonable to use the likes number as the natural measurement of comment quality. Based on this observation, we annotate the comments whose number of likes is more than 10 as high-quality comments.

Table~\ref{table-statics} presents some statistics of the dataset. We perform word segmentation for each sentence using a popular Chinese word segmentation toolkit\footnote{\url{https://github.com/fxsjy/jieba}}. It shows that the average number of words in one comment is 18.37, which is close to the title (16.64). But the vocabulary size of comment (53,916) is much larger than title (36,378). The reason is the expression in the user-generated comments are more informal and diverse.

As shown in Table~\ref{table-statics}, the average length of the news body is 326.17, which is too long. The abstract contains the main idea of the news, so we use the abstract instead of the news body to capture the content information.
We exclude the samples whose the length of title, abstract or comment is smaller than 5 or larger than 200. 
%

\begin{table}[t]
\centering
\small

\begin{tabular}{l|llll}
\hline
\bf Class      &  \bf Train   & \bf Valid  & \bf Test   & \bf Total   \\ \hline
Low   & 165,423 & 4,287  & 4,772  & 174,482 \\
High  & 197,331 & 5,713  & 5,228  & 208,272 \\ \hline
Total & 362,754 & 10,000 & 10,000 & 382,754 \\ \hline
\end{tabular}
\caption{Numbers of examples of different classes (low quality comment and high quality comment) in different sets.}
\label{tab:example number}
\vspace{-0.1in}
\end{table}

We divide the dataset into training, validation and test sets. Both the number of samples in validation set and test set are 10,000, and the number of samples in train set is 362,754. The numbers of examples of different classes in different sets in shown in Table~\ref{tab:example number}. 


\section{Proposed Model}

\begin{figure*}[t]
\centering
\includegraphics[scale=0.5]{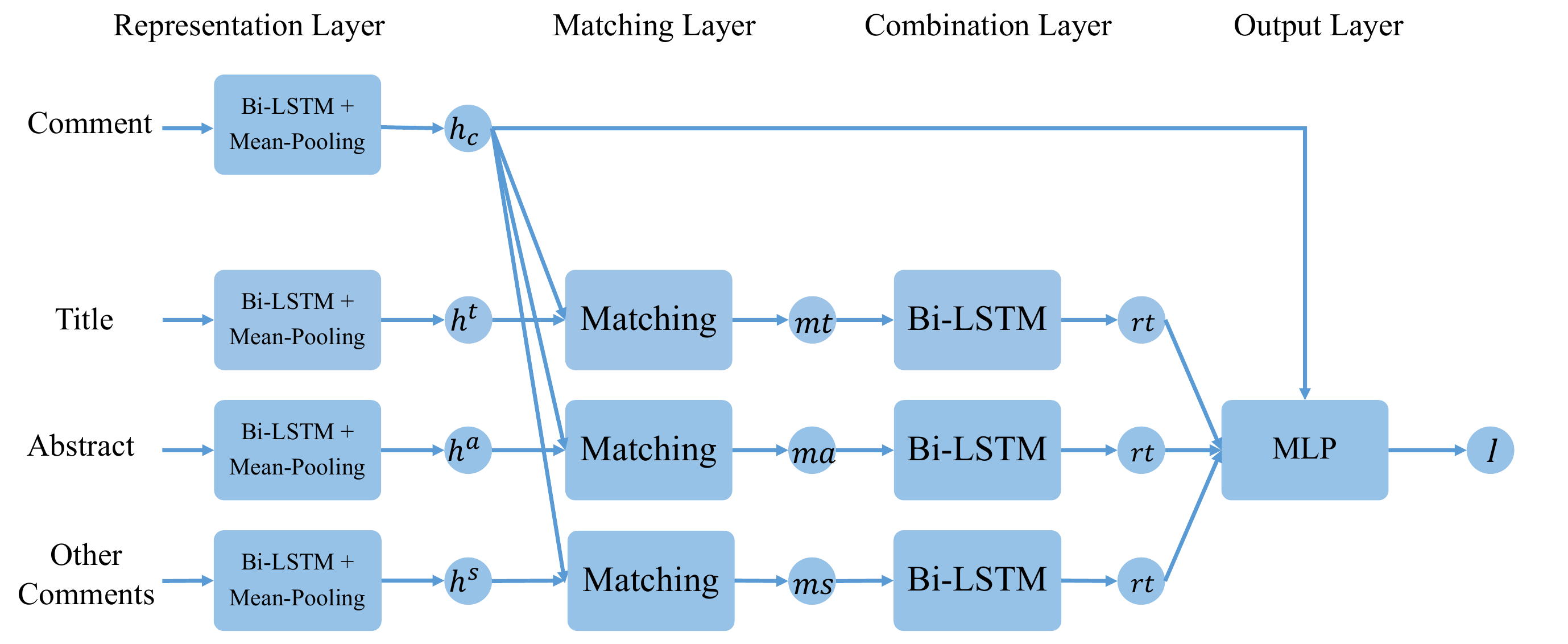}
\caption{The overview of the proposed MTM model.}
\label{fig:overview}
\vspace{-0.1in}
\end{figure*}

\subsection{Problem Formulation}
\label{problem}

Here, we give the notations and the formulation of the task. Suppose we have a set of $N$ example in dataset $\{x_1, x_2, \cdots, x_N\}$, and each example contains a title, an abstract, a comment,and several surrounding comments: $\bm{x} = \{\bm{t},\bm{a}, \bm{c}, \bm{s}$\}. The $\bm{s}$ denotes the context comments. 
Each comment has a label $l$ of whether the comment is high-quality or low-quality. Our goal is assigning the quality label for each upcoming comment. 

\subsection{Overview}
\label{overview}

In order to predict the quality label $l$, the proposed MTM (Multi-Target Text Matching) model estimates the probability distribution $P(l|\bm{x})=P(l|\bm{c},\bm{t}, \bm{a}, \bm{s})$. In our model, the quality of a news comment can be measured by means of three aspects: informativeness, consistency, and novelty. The informativeness is assessed by the comment itself. The consistency is evaluated by referring to the title and the abstract. And the novelty is assessed by comparing the comment with the surrounding comments. Our model takes consideration of these aspects and gives a general justification to the quality of the comment.

More specifically, our model first represents the comments, titles, and the abstract into vectors with the Bi-LSTM~\cite{DBLP:conf/icassp/GravesMH13}. Then the vectors are fed into a mean-pooling layer, and becomes text-level representations. After that, the representations of the comments are matched with the titles, abstracts, and the surrounding comments respectively. The combination layer is used to combine these three aspects, and the output layer finally predicts the quality label. The overview of the proposed model can be found in Figure~\ref{fig:overview}.

\subsection{Multi-Target Text Matching Model}
\label{layers}


We now give a detailed explanation of each component. Our model consists of the following four layers:

\noindent{\bf 1. Representation Layer:} The representation layer is to represent the comments, titles, and abstracts with dense vectors. It first transforms the words into word vectors $\bm{e}=\{e_1, e_2,\cdots,e_L\}$ ($L$ denotes the number of words). Then, the word vectors are fed into a Bi-LSTM to obtain the forward context representation and the backward context representation. The representations of the comment can be written as:
\begin{align}
\overrightarrow{\bm{s}}_i^c & = \overrightarrow{\rm LSTM}(\overrightarrow{\bm{s}}_{i-1}^c, \bm{e}_i) \label{equ1} \\
\overleftarrow{\bm{s}}_i^c & = \overleftarrow{\rm LSTM}(\overleftarrow{\bm{s}}_{i+1}^c, \bm{e}_i) \label{equ2}
\end{align}
where $i = 1,2,...,L$. The representations of the titles $[\overrightarrow{s}_{i}^{t}, \overleftarrow{s}_{i}^{t}]$ and the abstracts $[\overrightarrow{s}_{i}^{a}, \overleftarrow{s}_{i}^{a}]$ can be obtained in the similar way.


After getting the word-level representations, we use a mean-pooling layer to catch the n-gram information. We apply the overlapping mean-pooling layer to the hidden states in every time-step of Bi-LSTM. We calculate the average of the adjacent $ps$ hidden states and the stride is 1. The size of the mean-pooling $ps$ is a hyper-parameter. The experimental results show that this is helpful to improve the performance of the model. We also show formulas for the comment $\bm{c}$.
\begin{align}
\overrightarrow{\boldsymbol{h}}_{i}^{c} = \frac{\sum_{k=0}^{ps-1}\overrightarrow{\boldsymbol{s}}_{i+k}^{c}}{ps} \\
\overleftarrow{\boldsymbol{h}}_{i}^{c} = \frac{\sum_{k=0}^{ps-1}\overleftarrow{\boldsymbol{s}}_{i-k}^{c}}{ps}  
\end{align}
where $i = 1,2,...,L-ps+1$ . The similar computation is performed to obtain the representations of titles $[\overrightarrow{h}_{i}^{t}, \overleftarrow{h}_{i}^{t}]$ and abstracts $[\overrightarrow{h}_{i}^{a}, \overleftarrow{h}_{i}^{a}]$.

\noindent{\bf 2. Matching Layer:} The matching layer uses attention mechanism to measure the similarity between the comment and the title or the abstract. Besides, it measures the dissimilarity between the comment and the surrounding comments to assess the novelty. As is shown in Figure~\ref{fig:overview}, 
for each hidden state in the comment representations, all hidden states in the context representations (title,abstract and surrounding comments) will be matched independently.

We now take the matching between the title $\bm{t}$ and the comment $\bm{c}$ as the example. First, we calculate the attention weights of both directions for the $i$-th hidden state of the comment:
\begin{align}
\overrightarrow{\boldsymbol{ \alpha }}_{i,j} =\cos(\overrightarrow{\boldsymbol{h}}_{i}^{c},\overrightarrow{\boldsymbol{h}}_{j}^{t}) \\
\overleftarrow{\boldsymbol{ \alpha }}_{i,j} =\cos(\overleftarrow{\boldsymbol{h}}_{i}^{c},\overleftarrow{\boldsymbol{h}}_{j}^{t})
\end{align}
where $i = 1,2,...,L_c'$ and $j = 1,2,...,L_t'$. ($L_c'$ and $L_t'$ denote the number of hidden states of comment and title's hidden states after pooling, respectively.) 
Then, we take $\overrightarrow{\boldsymbol{\alpha }}_{i,j}$($\overleftarrow{\boldsymbol{ \alpha }}_{i,j} $) as the weight of $\overrightarrow{\boldsymbol{h}}_{j}^{t}$($\overleftarrow{\boldsymbol{h}}_{j}^{t}$), and calculate an attentive vector for the entire title $\bm{t}$ by weighted summing all the $\overrightarrow{\boldsymbol{h}}_{j}^{t}$($\overleftarrow{\boldsymbol{h}}_{j}^{t}$):
\begin{equation}
\overrightarrow{\boldsymbol{h}}_{i}^{t_{sum}} = \frac{\sum_{j=1}^{L_t'}\overrightarrow{\boldsymbol{ \alpha }}_{i,j}\ast  \overrightarrow{\boldsymbol{h}}_{j}^{t}}{\sum_{j=1}^{L_t'}\overrightarrow{\boldsymbol{ \alpha }}_{i,j}} 
\end{equation}
\begin{equation}
\overleftarrow{\boldsymbol{h}}_{i}^{t_{sum}} = \frac{\sum_{j=1}^{L_t'}\overleftarrow{\boldsymbol{ \alpha }}_{i,j}\ast  \overleftarrow{\boldsymbol{h}}_{j}^{t}}{\sum_{j=1}^{L_t'}\overleftarrow{\boldsymbol{ \alpha }}_{i,j}} 
\end{equation}
where $i = 1,2,...,L_c'$.

After getting the weighted-sum vectors, we perform the matching operation:
\begin{align}
\overrightarrow{\boldsymbol{ mt }}_{i}^{k} =f_{m}(\overrightarrow{\boldsymbol{h}}_{i}^{c},\overrightarrow{\boldsymbol{h}}_{i}^{t_{sum}},W^{k}  ) \\
\overleftarrow{\boldsymbol{ mt }}_{i}^{k} =f_{m}(\overleftarrow{\boldsymbol{h}}_{i}^{c},\overleftarrow{\boldsymbol{h}}_{i}^{t_{sum}}, W^{k}  )
\end{align}
where $i = 1,2,...,L_c'$ and $k = 1,2,...,p$, $p$ is the number of perspectives~\cite{DBLP:conf/ijcai/WangHF17}. And the $f_{m}$ is defined in the following way:
\begin{equation}
f_m(v_1, v_2, W) = \cos(\boldsymbol{v_1}\circ W,\boldsymbol{v_2}\circ W) 
\end{equation}
where the $\circ$ is the element-wise multiplication and the $W$ is the parameter matrix.

Finally, we get the matching vectors for the title from different perspectives. The matching vectors for abstract and comment can be obtained in a similar way.
\begin{align}
\bm{\overrightarrow{mt_i}} &= [\overrightarrow{mt}_i^{1},\overrightarrow{mt}_i^{2}, ..., \overrightarrow{mt}_i^{p}]  \\
\bm{\overleftarrow{mt_i}} &= [\overleftarrow{mt}_i^{1},\overleftarrow{mt}_i^{2}, ..., \overleftarrow{mt}_i^{p}]
\end{align}
where $i = 1,2,...,L_c'$. Now we get the matching result vectors (in two directions) between title and comment: $\overrightarrow{\bm{mt_i}}(\overleftarrow{\bm{mt_i}})$. We can also get other matching result $\overrightarrow{\bm{ma_i}}(\overleftarrow{\bm{ma_i}})$, $\overrightarrow{\bm{ms_i}}(\overleftarrow{\bm{ms_i}})$ by the same way. 
 
The final matching result is obtained by connecting the matching result from two directions.
 \begin{align}
\bm{mt_i} &= [\bm{\overrightarrow{mt_i}},\bm{\overleftarrow{mt_i}}] \\
\bm{ma_i} &= [\bm{\overrightarrow{ma_i}},\bm{\overleftarrow{ma_i}}] \\
\bm{ms_i} &= [\bm{\overrightarrow{ms_i}},\bm{\overleftarrow{ms_i}}]
\end{align}
where $i = 1,2,...,L_c'$.
Here the $\bm{mt_i}(\bm{ma_i},\bm{mc_i})$ is the matching result for one time-step, so we connect all the time-steps' results and get the matching results $\bm{mt}(\bm{ma},\bm{mc})$ for the whole sentences.

\noindent{\bf 3. Combination Layer:} The combination layer is to combine different components of matching vectors into a vector for prediction. In our model, the quality of news comments can be measured from three aspects: informativeness, consistency and novelty. The informativeness is directly represented by the mean-pooling of comment's representation.
\begin{equation}
\bm{R}_{info} = \frac{\sum_{i=1}^{L}\boldsymbol{s}_{i}^{c}}{L} 
\end{equation}
where $i = 1,2,...,L$, and $L$ denotes the number of words in comment. 

In the previous layer, we get the matching result: $\bm{mt_i}$ and $\bm{ma_i}$. Here we use another Bi-LSTM to process the matching result:
\begin{align}
\overrightarrow{\bm{rt}}_i & = \overrightarrow{\rm LSTM}(\overrightarrow{\bm{rt}}_{i-1}, \bm{mt}_i) 
\label{lstm2-1}
\\
\overleftarrow{\bm{rt}}_i & = \overleftarrow{\rm LSTM}(\overleftarrow{\bm{rt}}_{i+1}, \bm{mt}_i) 
\label{lstm2-2}
\end{align}
where $i = 1,2,...,L'_c$.
After this, we choose the last time-step of both directions to form the vector for prediction.
\begin{align}
\bm{rt} = [\overrightarrow{\bm{rt}}_{L'_c},\overleftarrow{\bm{rt}}_{1}]
\end{align}
Similarly, we get $\bm{ra}$ and $\bm{rc}$. The consistency is measured by the matching result between comment and title(abstract). And the novelty is directly measured by the matching result between comment and surrounding comments. 
\begin{center}
\begin{align}
\bm{R}_{cons} &= [\bm{rt},\bm{ra}] \\
\bm{R}_{nove} &= \bm{rc}
\end{align}
\end{center}
Then we just connect all this three parts and get the final vector for prediction.
\begin{equation}
\bm{R} = [\bm{R}_{info} ,\bm{R}_{cons},\bm{R}_{nove}]
\end{equation}

\noindent{\bf 4. Output Layer:} The out layer is to evaluate the probability distribution $P(l|\bm{t,a,c,s})$ and output the prediction of comment label . In this layer, we simply use three layer feed-forward neural network to predict the result. 
\begin{equation}
p(l|\bm{c,t,a,s})=softmax(\bm{W_oR}+\bm{b_o})
\end{equation}
where $\bm{W_o, b_o}$ is trainable parameters.

\section{Experiments}

In this section, we perform experiments to evaluate our model on the Toutiao Comment Dataset.

\subsection{Experimental Details}
We adopt the accuracy and $\rm F_1$ score as our main evaluation metrics, which is used in various classification tasks. In addition, precision and recall scores are also reported to assist the analysis.
The word embedding with 200 dimension is initialized using word2vec~\cite{DBLP:conf/nips/MikolovSCCD13}. The hidden size of Bi-LSTM is 100, and the number of layer is 1. 
We use the Adam~\cite{KingmaBa2014} optimizer with the initial learning rate $\alpha = 0.001$. Besides, the dropout regularization~\cite{dropout} with the dropout probability $p=0.2$ is used to reduce overfitting. 

\subsection{Baselines}
We compare our model with the following baselines:
\begin{itemize}
    \item \noindent\textbf{Traditional machine learning methods:}
    We choose several traditional machine learning classifiers, including SVM, MultinomialNB, BernoulliNB, LogisticRegression, DecisionTree, RandomForest, and AdaBoost. 
    \item \noindent\textbf{Siamese-CNN:}
    We use the Siamese framework and use CNN to get the text representation. The kernel size is [3,4,5] and the kernel number is 50. 
    \item \noindent\textbf{Siamese-LSTM:}
    We use the Siamese framework and use Bi-LSTM to obtain the text representation. The hidden dimension of LSTM is 100.
    \item \noindent\textbf{BIMPM~\cite{DBLP:conf/ijcai/WangHF17}:}
    BIMPM is a popular model to predict a label with matching two sentences. We use BIMPM model as a baseline to match the comments and the other contexts(as a whole).
\end{itemize}

\subsection{Results}
Here we compare our proposed model with the baselines. Table~\ref{tab1} reports the experiment results of various models.

\begin{table}[tb]
		\centering
        \small
		\begin{tabular}{ l | c c c c }
		\hline
		\multicolumn{1}{ l |}{\textbf{Models}} &
		\multicolumn{1}{c}{\textbf{Acc(\%)}} & 
		\multicolumn{1}{c}{\textbf{P(\%)}} &  
        \multicolumn{1}{c}{\textbf{R(\%)}} &
		\multicolumn{1}{c }{\textbf{F1(\%)}}   \\ \hline
       SVM & 61.59 & 71.20 & 72.37 & 71.18 \\
       LR & 63.59 & 70.80 & 78.40 & 74.41 \\
       Decision Tree & 57.48 & 70.74 & 63.13 & 66.72 \\
       MNB & 62.63 & 69.63 & 79.17 & 74.09 \\
       BNB & 59.80 & \textbf{73.36} & 62.97 & 67.90 \\
       Random Forest & 60.99 & 71.92 & 69.26 & 70.57 \\
       AdaBoost & 60.48 & 71.85 & 68.17 & 69.96 \\ \hline
       Siamese-CNN & 65.68 & 67.07 & 87.53 & 75.94 \\
       Siamese-LSTM & 66.17 & 67.73 & 86.59 & 76.00 \\
       BIMPM & 67.48 & 72.86 & 82.55 & 77.40 \\ \hline
       \textbf{MTM (this work)} & \textbf{70.75} & 72.66 & \textbf{90.83} & \textbf{80.73} \\ \hline
		\end{tabular}
		\caption{Comparison between our proposed model and the baselines on the test set (Acc, P, R, and F1 denote accuracy, precision, recall and ${\rm F_1}$ score, respectively).}
		\label{tab1}
        \vspace{-0.1in}
\end{table}





\begin{table}[t]
\centering
\small
\begin{tabular}{c|cccc}
\hline
\textbf{Correlation} & \textbf{Info}      & \textbf{Cons} & \textbf{Nove} & \textbf{Total} \\ \hline
Spearman    & \textbf{0.740} & 0.574 & 0.610 & 0.689 \\
Pearson     & \textbf{0.745} & 0.544 & 0.608 & 0.704 \\ \hline
\end{tabular}
\caption{The correlation analysis between human scoring and our model's scoring in different metrics. All the correlation is significant with $p<0.05$ (\textbf{Info} denotes informativeness, \textbf{Cons} denotes consistency, and \textbf{Nove} denotes novelty).}
\label{table:human2}
\vspace{-0.1in}
\end{table}

\begin{table*}[tb]
		\centering
        \small
    	\setlength{\tabcolsep}{10.0pt}
		\begin{tabular}{ l | c c c c }
		\hline
		\multicolumn{1}{ l |}{\textbf{Models}} &
		\multicolumn{1}{c}{\textbf{Acc(\%)}} & 
		\multicolumn{1}{c}{\textbf{P(\%)}} & 
        \multicolumn{1}{c}{\textbf{R(\%)}} & 
		\multicolumn{1}{c }{\textbf{F1(\%)}}   \\ \hline
      Full Model & 70.75 & 72.66 & 90.83 &80.73 \\ \hline
      \emph{w/o title} & 70.10($\downarrow 0.65$) & 73.34($\uparrow 0.68$) & 87.47($\downarrow 3.36$) & 79.79($\downarrow 0.94$) \\
      \emph{w/o abstract} & 69.82($\downarrow 0.93$) & 74.30($\uparrow 1.64$) & 84.48($\downarrow 6.35$) & 79.07($\downarrow 1.66$) \\
      \emph{w/o surrounding comments} & 69.16($\downarrow 1.59$) & 71.78($\downarrow 0.88$) & 89.46($\downarrow 1.37$) & 79.56($\downarrow 1.17$) \\
      \hline
	\end{tabular}
	\caption{Ablation Study. Performance on the test set when removing different parts of text.}
	\label{tab2}
    \vspace{-0.1in}
\end{table*}

\begin{table}[tb]
		\centering
        \footnotesize
		\begin{tabular}{ c | c c c c }
		\hline
		\multicolumn{1}{ c |}{\textbf{Number}} &
		\multicolumn{1}{c}{\textbf{Acc(\%)}} & 
		\multicolumn{1}{c}{\textbf{P(\%)}} &  
        \multicolumn{1}{c}{\textbf{R(\%)}} &
		\multicolumn{1}{c }{\textbf{F1(\%)}}   \\ \hline
        0 & 69.16 & 71.78 & 89.46 & 79.56 \\
        1 & 68.22 & 72.03 & 86.48 & 78.59 \\
        3 & 69.53 & \textbf{73.45}& 85.89 & 79.18 \\
        5 & \textbf{70.75} & 72.66& \textbf{90.83} & \textbf{80.73} \\  \hline
		\end{tabular}
		\caption{Performance on the test set with different number of surrounding comments. }
		\label{tab3}
        \vspace{-0.1in}
\end{table}

\begin{table}[tb]
		\centering
        \footnotesize
		\begin{tabular}{ c | c c c c }
		\hline
		\multicolumn{1}{ c |}{\textbf{Pooling Size}} &
		\multicolumn{1}{c}{\textbf{Acc(\%)}} & 
		\multicolumn{1}{c}{\textbf{P(\%)}} &  
        \multicolumn{1}{c}{\textbf{R(\%)}} &
		\multicolumn{1}{c }{\textbf{F1(\%)}}   \\ \hline
        1 & 69.91 & 71.38 & 86.24 & 78.11 \\
        2 & 70.45 & 71.71 & 87.68 & 78.89 \\
        3 & 70.68 & \textbf{72.89} & 89.48 & 80.34 \\
        4 & \textbf{70.75} & 72.66 & \textbf{90.83} & \textbf{80.73} \\  \hline
		\end{tabular}
		\caption{Performance on the test set with different pooling size.}
		\label{tab4}
        \vspace{-0.1in}
\end{table}

As is shown in Table~\ref{tab1}, our proposed MTM model achieves the best performance in the main evaluation metrics. Among all the traditional machine learning models, the logistic regression method achieves the highest accuracy and $\rm{F_1}$ score. However, our MTM model achieves improvements of 7.16\% accuracy and 6.32\%  $\rm F_1$ score over the logistic regression model. In addition, our proposed MTM model is also able to outperform other existing neural network models by a large margin. For instance, our MTM model achieves improvements of 3.27\% accuracy and 3.33\%  $\rm F_1$ score over the popular BIMPM model, which shows that the multi-target matching mechanism can effectively improve the performance of model classification. MTM model can match the comment and multi-kind of auxiliary information in a more effective way. Therefore, our proposed model is capable to learn better representation for classification.

\subsection{Human Evaluation}

\begin{figure}[t]
\centering
\includegraphics[width=1.0\linewidth]{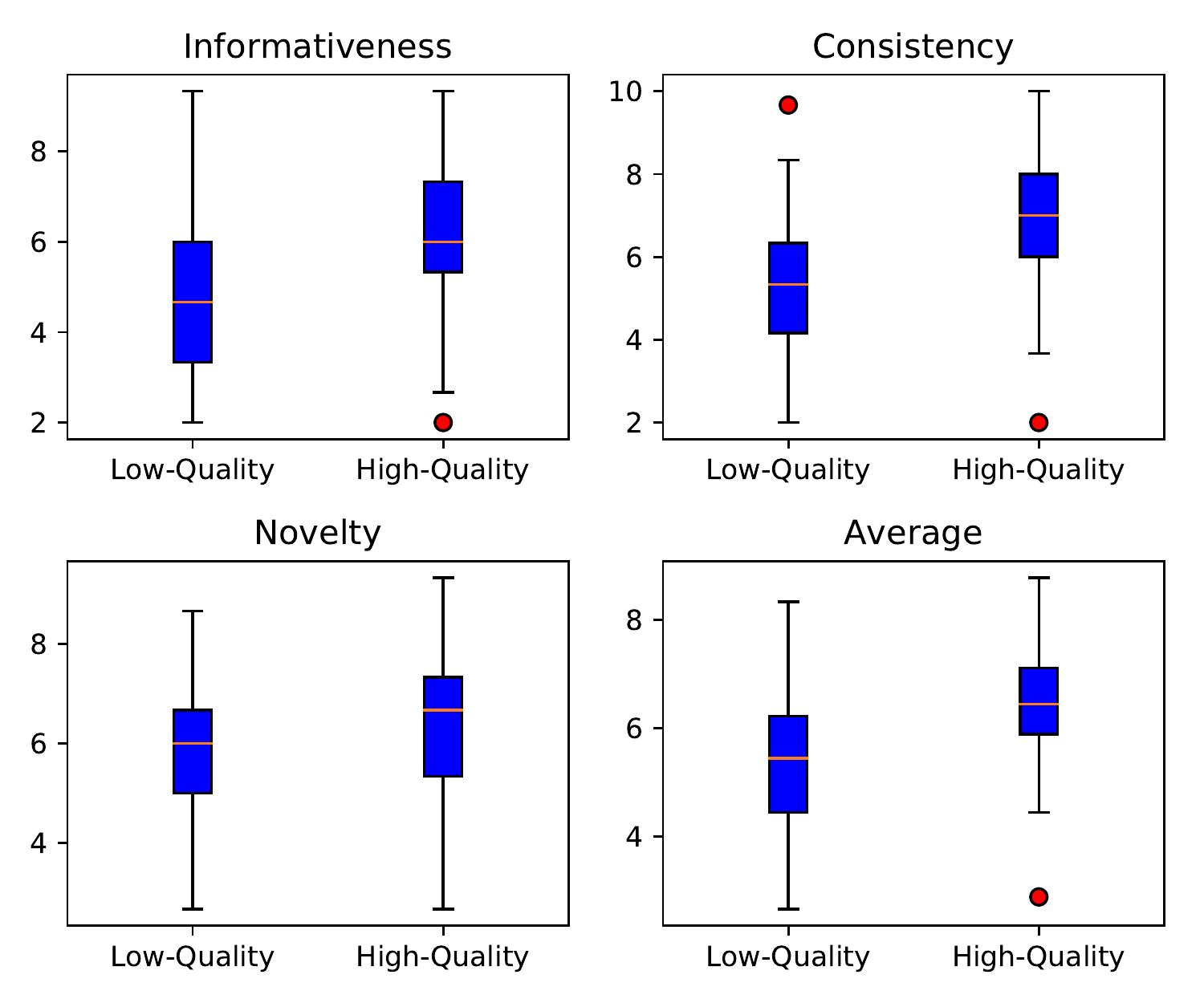}
\vspace{-2em}
\caption{The box-plot for three metrics of different quality comments. Red dots represent outliers.}
\label{fig:box-plot}
\vspace{-0.2in}
\end{figure}

In this paper, the quality of comments is measured in three metrics: informativeness, consistence and novelty. Here come \textbf{two important questions}: can these metrics measure the quality of comment well? and does our model realize the measurement of the metrics successfully? Since these metrics are subjective, we use human evaluation and statistical analysis to analyze two questions. 

We randomly select 120 examples from the test set, and we assign three annotators to evaluate the comments independently. Each comment is evaluated with a 10-point scale in three metrics: the informativeness of a comment itself, the consistency between the comment and the news, and the novelty of comments compared with the surrounding comments. We average three annotators' scores for each metric to obtain the human scores. 

To answer the \textbf{first question}, we analyze the relationship between the human scores and the quality label of comment. We conduct the independent sample $t-$test for annotators' score based on comment's quality label. The results show that there are significant differences ($p<0.05$) of the mean value of four human scores between high-quality class and low-quality class. For better visualizing the significance, we use the box-plot to present the difference between high-quality class and low-quality class, as shown in Figure~\ref{fig:box-plot}. It concludes that the metrics we use in this work can measure the comment quality well.

To analyze the \textbf{second question}, we obtain our model's scores on three metrics by removing different parts of text. We scale the output probabilities to [0, 10] so that it is comparable to the human scores. We conduct the correlation analyze between our model's scores and the human scores. We calculate the Pearson correlation coefficient and Spearman correlation coefficient for all the three scores as well as the total score. The result is shown in Table~\ref{table:human2}. We find that all these scores are significantly correlated ($p<0.05$) between human and model's results. It concludes that our model realize the measurement of three metrics successfully. Besides, among these three scores, the correlation coefficient of informativeness is highest, which indicates that the informativeness is more important in our model.





\subsection{Impact of Different Parts of Text}
Here we explore the impact of model inputs to its performance by removing different parts of text. The related experiment result is shown in Table~\ref{tab2}. 
%
As is shown in Table~\ref{tab2}, the performance of the model shows different degrees of decline when we remove different context text. This shows that each input context is helpful for the classification and there are differences in the contribution of different context to the performance of the model. There is the smallest decline in the model performance when removing the title of news. It is reasonable because the news title tends to contain limited information.


\subsection{Impact of the Surrounding Comments}
In order to assess the novelty of a given comment, we also use the surrounding comments about this news as input text. Here the impact of the number of the surrounding comments on the model performance is further analyzed and the related experiment result is shown in Table~\ref{tab3}.

According to Table \ref{tab3}, we find that with the increase of the number of the surrounding comments, the model performs better, which shows that the surrounding comments are of great help for classification. The proposed model can refer to the surrounding comments for analyzing the novelty of a given comment. The larger the number of surrounding comments, the more input information can be enriched, leading to a more accurate assessment of novelty. However, we find that when only one surrounding comment is used, the performance of the model turns worse compared to using no surrounding comment. The reason is that the model suffers a large variance in the case where there is only one comment, making the novelty score inaccurately evaluated.


\subsection{Impact of the Mean-Pooling Operation}
In our model, the mean-pooling operation is used to improve the local information and is helpful to the performance. The reason is that richer local features similar to $n$-grams are obtained by means of the mean-pooling operation. Here we conduct further analysis on the pooling-size of mean-pooling and the result is shown in Table~\ref{tab3}.

As is shown in Table~\ref{tab3}, as the pooling-size increases, the performance of the model continues to improve in all main matrices. This shows that the model can get better representations through the mean-pooling operation, leading to better performance. Through the mean-pooling layer, several adjacent hidden state vectors are merged to output more abundant local features. Therefore, the proposed model is capable of capturing both the global and local textual semantics to form better representations for classification. 

\begin{figure*}[t]
\centering
\includegraphics[scale=0.43]{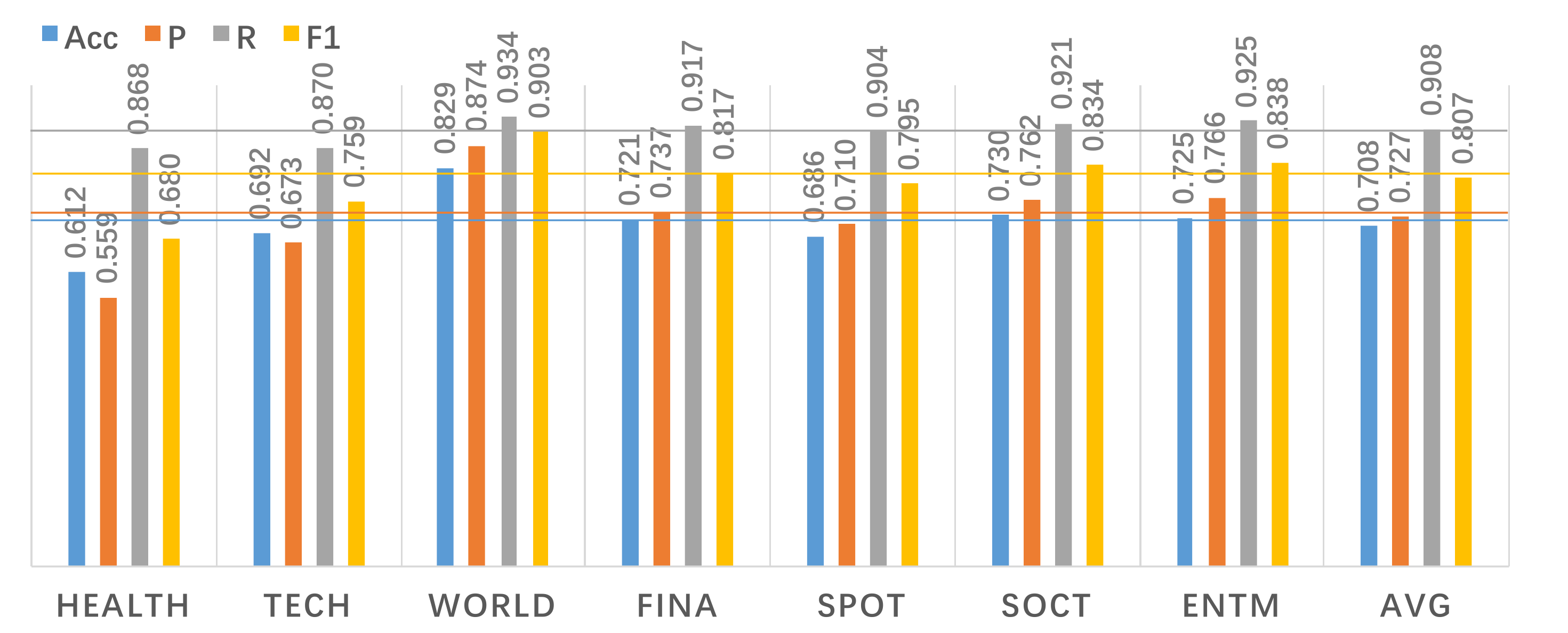}
\caption{Results of different types of news. The types from left to right are: health, technology, world, finance, sports, society, entertainment and average result.}
\label{fig:type}
\vspace{-0.1in}
\end{figure*}

\begin{figure}[t]
\centering
\includegraphics[scale=0.2]{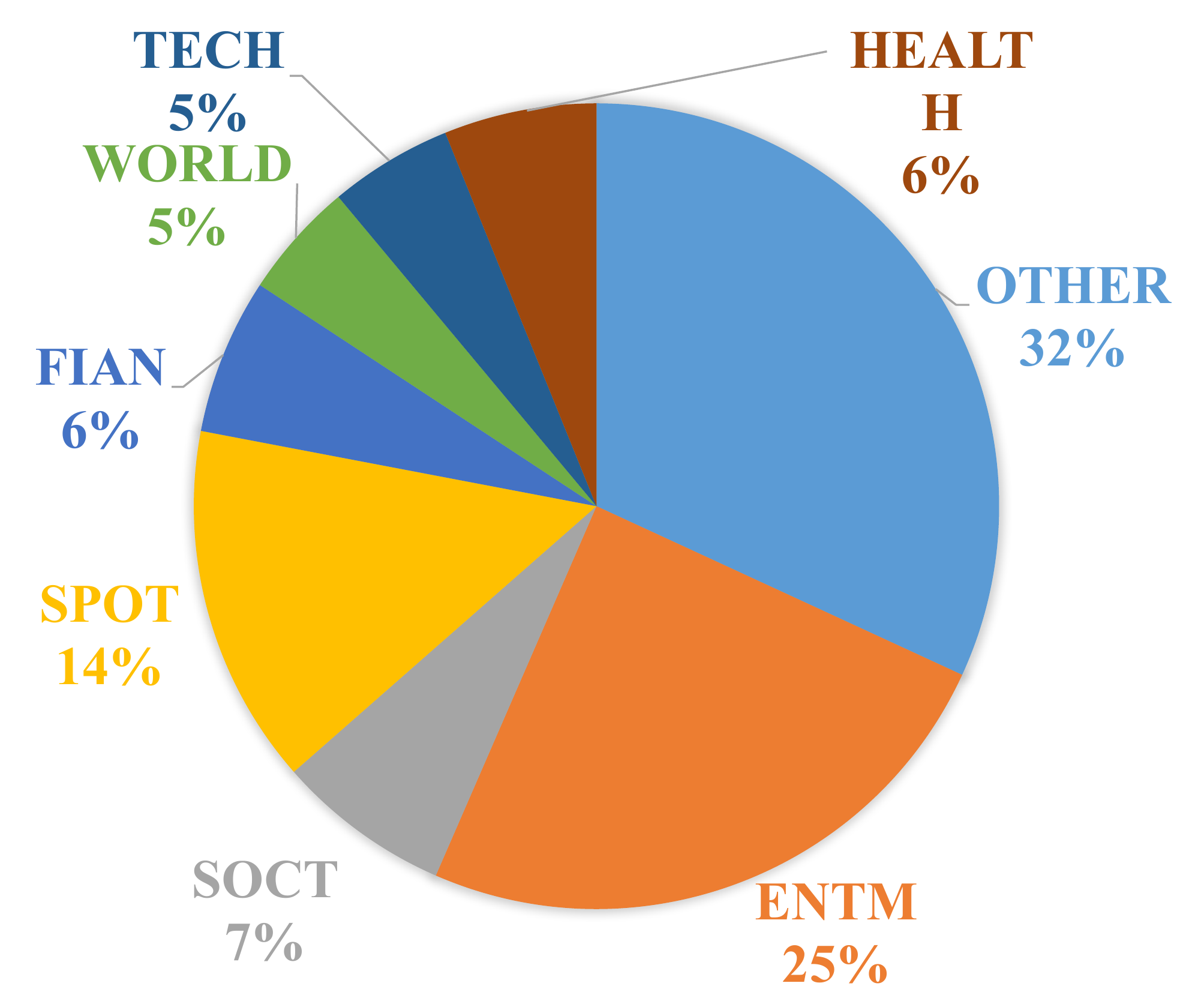}
\caption{Pie Chart of the number of news in each type in train set.}
\label{fig:type num}
\vspace{-0.18in}
\end{figure}

\subsection{Error Analysis}

We find that there are significant differences in the performance of the model on different type news. In order to explore the impact of the news type, we select seven different news types in our test set, and each type has at least several hundred samples. The performance of the model on these seven different types of news is shown in Figure~\ref{fig:type}.

According to Figure~\ref{fig:type}, we find that the performance of the model on world news is obviously better than average (accuracy 82.9\% vs 70.8$\%$, $\rm F_1$ score 90.3\% vs 80.7\%). However, the model performance on the health news is obviously worse than average (accuracy 61.2\% vs 70.8\%, $\rm F_1$ score 68.0\% vs 80.7\%). 

To analyze this phenomenon, we first count the number of news in each type in our training set. The result is shown in Figure~\ref{fig:type num}. And we can see that the number of world news is close to health news. At the same time, the number of entertainment news is much larger than finance news, but they have similar result in test. So we can conclude that the number of examples in train set has little influence on the result.

Then why the results can be so different in world news and health news? We think it can be explained that the world news contains less professional knowledge. So it is easy to arouse the user's resonance to give reasonable feedback. At the same time, less professional knowledge makes it easy to capture the relevant semantic features, leading that the proposed model can learn an effective pattern to perform classification. However, there are a large amount of expertise in the health news, leading to sparse data. Therefore, it is difficult for the model to learn a unified pattern for classification, resulting in poor performance.


\section{Related Work}
There have been some studies about news comments. \citet{DBLP:conf/coling/MaW10} try to extract opinion target from news comments. Their method use global information in news articles as well as contextual information in adjacent sentences of comments. \citet{DBLP:conf/acllaw/NapolesTPRP17} try to identify “good” conversations that occur online. They build the Yahoo News comment threads Dataset and try to find Engaging, Respectful, and/or Informative Conversations. This dataset handles a thread of comment as a whole. \citet{DBLP:conf/chi/ParkSDE16} develop a system, CommentIQ, which supports comment moderators in interactively identifying high quality comments. 
\citet{DBLP:conf/emnlp/KolhatkarT17} also proposes a model to classifier the comments, and they focuses on the constructive comments, which is different from ours.

Many tasks can be formulated as the text matching problem, such as question answering~\cite{qa1,qa2} and dialogue generation~\cite{dia1}. Siamese framework~\citep{DBLP:journals/ijprai/BromleyBBGLMSS93} is a classical method to deal with the Natural Language Sentence Matching(NLSM) task. 
\citet{Bian0YCL17} proposed Matching-Aggregation framework to overcome this problem. 

\citet{DBLP:conf/nips/HuLLC14} proposed ARC-II model, which connects the n-gram of the two sentences and builds a 2D matrix first and then conduct matching. \citet{DBLP:conf/aaai/PangLGXWC16} proposed Match-Pyramid model, which transfers the text matching to image recognition by calculate the similarity matrix first.  \citet{yang2016anmm-attention} find that attention architecture is helpful for the matching result. \citet{DBLP:conf/coling/SugguGCS16-fusion} propose a fusion model that uses deep-learning features and artificial features at the same time. \citet{DBLP:conf/ijcai/WangHF17} propose BIMPM model, and they match the two sentences in two directions and multi-views on each hidden state of Bi-LSTM. \citet{DBLP:conf/www/LiuZHNLX18-hierarchical} propose a hierarchical model for sentences matching. 

There is previous work regarding rating the academic paper~\cite{paperrating}, while this work is about rating the news comments, which is different from them.



\section{Conclusions}

In this work, we propose the task of high-quality comment identification, and construct a large-scale annotated dataset. 
We analyze the dataset, and find the quality of comments can be measured in three aspects: informativeness, consistency, and novelty.
In order to measure three aspects above automatically, we propose Multi-Target Text Matching model. Experimental results show that our model's scoring are highly correlated with human scoring in three aspects. Besides, our model outperforms various baselines by a large margin. 



\bibliography{emnlp2018}
\bibliographystyle{acl_natbib_nourl}

\end{document}